\documentclass{article}
\usepackage{microtype}
\usepackage{graphicx}
\usepackage{subcaption}
\usepackage{booktabs} %

\usepackage{hyperref}

\usepackage[preprint]{meta/icml2026}

\usepackage{amsmath}
\usepackage{amssymb}
\usepackage{mathtools}
\usepackage{amsthm}

\usepackage{enumitem}

\begin{document}
\twocolumn[
  \icmltitle{Domain-Adaptable Reinforcement Learning for Code Generation with Dense~Rewards}

  \icmlsetsymbol{equal}{*}

  \begin{icmlauthorlist}
    \icmlauthor{Erfan Aghadavoodi Jolfaei}{tuda}
    \icmlauthor{Daniel Maninger}{tuda,hai}
    \icmlauthor{Abhinav Anand}{tuda}
    \icmlauthor{Mert Tiftikci}{tuda}
    \icmlauthor{Mira Mezini}{tuda,hai,athene}
  \end{icmlauthorlist}

  \icmlaffiliation{tuda}{Technische Universität Darmstadt, Darmstadt, Germany}
  \icmlaffiliation{hai}{Hessian Center for Artificial Intelligence (hessian.AI), Darmstadt, Germany}
  \icmlaffiliation{athene}{National Research Center for Applied Cybersecurity ATHENE, Darmstadt, Germany}

  \icmlcorrespondingauthor{Erfan Aghadavoodi Jolfaei}{erfan.aghadavoodi@gmail.com}

  \icmlkeywords{LLMs, RL, PPO, Code Generation, Robotics}

  \vskip 0.3in
]

\printAffiliationsAndNotice{}  %

\begin{abstract}

Large language models show strong potential for automated code generation, but lack guarantees for correctness, quality, safety, and domain-specific constraints. For instance in robotics, where code generation is increasingly being used for planning and executing actions, awareness of the environment and physical constraints is critical. To facilitate the adaption of code-generating LLMs to diverse requirements, including domain-specific ones, we present a reinforcement learning framework that fine-tunes pre-trained LLMs using proximal policy optimization. Our customizable execution-aware reward formula captures and optimizes syntax, functional correctness, code style, security, and simulator executability. A token-level reward mapping mechanism enables effective credit assignment from execution outcomes to generated tokens. The framework is evaluated on general-purpose code generation (MBPP/MBPP+) and robotic program synthesis (\textsc{RoboEval}). The results show substantial improvements in functional correctness and simulator executability, including an absolute pass@1 increase of 19\% on MBPP and a reduction in execution failures by 51\% on \textsc{RoboEval}. These findings demonstrate that structured reinforcement learning can effectively align language models to correct program generation and domain-specific requirements.
\end{abstract}

\begin{figure*}
\centering
\includegraphics[width=1.0\textwidth]{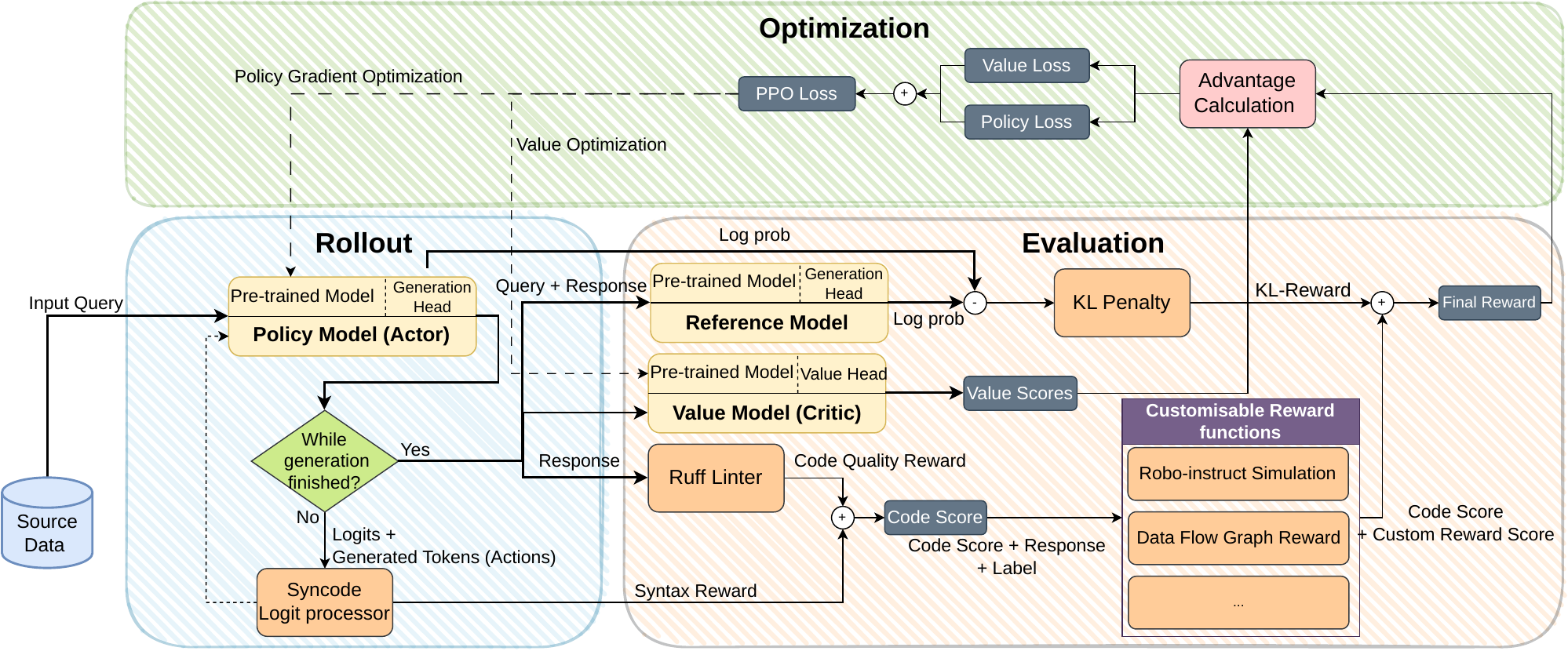}
\caption{Overview of the proposed fine-tuning framework. The process operates in a loop of \textbf{Rollout}, \textbf{Evaluation}, and \textbf{Optimization}.}
\label{fig:framework-structure}
\end{figure*}

\section{Introduction}

Large language models (LLMs) have attracted significant attention for their potential in automated code generation, promising to enhance developer productivity and accelerate software development~\cite{llms_software_engineering}. However, the code generated by these models can suffer from syntactic, functional, or non-functional flaws.

These limitations become even more critical when LLMs are used in domains that are subject to special requirements and constraints. In robotics, for instance, code generation enables robots to autonomously reason about tasks, plan high-level actions, and synthesize executable control programs~\cite{mobile_robots}. Deploying LLM-generated code in embodied robotic systems introduces stringent requirements: programs must be safe to execute on physical hardware, reliable under real-world uncertainty, and continuously responsive to a dynamic environment. In this setting, generated code must not only compile and pass basic unit tests but also functionally align with the specific domain and task requirements.

Reinforcement learning with verifiable reward (RLVR) has emerged as a promising approach to address limitations of code-generating LLMs by optimizing models using execution-based feedback~\cite{qwen_coder, deepseekcoder}. However, current implementations of RLVR for code depend on sparse sequence-level rewards, which penalizes the entire code sequence in case of error. Distillation-based methods provide more fine-grained feedback but are computationally inefficient and require access to a very large models~\cite{sdpo}.

In this work, we address these limitations by introducing a \emph{token-level reward formulation} for reinforcement learning (RL) in code generation. Instead of relying solely on sequence-level feedback, we transform multiple evaluation signals---including syntactic validity, static code analysis, and execution outcomes---into dense and informative token-level rewards. This enables fine-grained credit assignment, allowing the model to learn which parts of the generated code contribute to success or failure.

We integrate this formulation into a unified fine-tuning framework based on proximal policy optimization (PPO)~\cite{ppo_algo}. A pre-trained LLM acts as the policy (actor), while a secondary model serves as the critic to estimate expected rewards. The set of reward functions is customizable to support the adaption of LLMs to different domains.

In summary, the main \textbf{contributions} of this work are:

\begin{itemize}[noitemsep]
\item We introduce a \textbf{unified PPO-based fine-tuning framework} combining syntactic constraints, static analysis, execution results, and simulator feedback as rewards for program generation.

\item We propose a \textbf{dense token-level reward attribution mechanism} that transforms sparse sequence-level feedback into localized training signals.

\item We demonstrate the adaptability and efficacy of our framework in two different settings: general programming and robotics tasks. Our experiments show that compact language models can achieve \textbf{significant gains in correctness and executability}, respectively.
\end{itemize}

\section{Method}

Figure~\ref{fig:framework-structure} illustrates the proposed fine-tuning framework, which operates as an iterative PPO~\cite{ppo_algo} loop. The framework aligns the LLM (policy model) with multiple code-quality objectives, combining token-level and sequence-level rewards for comprehensive code generation improvement. An algorithmic breakdown of the whole process is provided in Appendix~\ref{sec:training-algorithm}.

\subsection{Training Phases}

The training process consists of a loop of three phases:

\paragraph{Rollout Phase: Token-wise Generation and Feedback}
An input prompt is fed into the policy model $\pi_\theta$, which then generates code one token at a time. During this process, a logits processor evaluates each token for syntactic correctness, providing immediate token-level feedback.

\paragraph{Evaluation Phase: Multi-Channel Reward Aggregation}
Once a full code sequence is generated, the framework computes a comprehensive final reward by aggregating various reward components including the KL divergence to a reference model $\pi_\text{ref}$ and a customizable set of code quality scores, which will be introduced in the next section. In addition, a value model $V_\phi$ computes a value score used for calculating the advantage in the optimization phase.

\paragraph{Optimization Phase: Learning from Feedback}
The final reward and the value score are used to calculate the advantage and loss, which are then used to update the policy and value model parameters.

\subsection{Reward Formulation}
\label{sec:composite-reward}

A key feature of our framework is the application of a \emph{multi-component reward}, which extends standard RLVR~\cite{qwen_coder, deepseekcoder} by simultaneously evaluating complementary aspects of code quality, enabling \emph{iterative, reward-driven refinement} across multiple dimensions. The training signal is derived from the weighted sum of four distinct components, each integrating a specific quality standard directly into the training loop:

\begin{itemize}[noitemsep]
    \item \textbf{Syntactic Correctness} ($R_{\text{sync}}$): A modified version of the \textsc{SynCode} constrained decoding algorithm~\cite{syncode} flags tokens that violate the language grammar.

    \item \textbf{Code Style and Vulnerabilities} ($R_{\text{lint}}$): The Ruff linter~\cite{ruff} detects syntax errors, common logical issues, and potential security vulnerabilities.

    \item \textbf{Distributional Regularization} ($R_{\text{KL}}$): The KL divergence from the reference model, used to stabilize policy updates and prevent catastrophic policy drift.

    \item \textbf{Optional Task-specific Rewards} ($R_{\text{opt}^i}$): Customizable reward functions that can be used to adapt the framework to different code generation settings. In our experiments, three task-specific rewards are implemented:
    \begin{itemize}[noitemsep]
        \item Pass@1 unit test results
        \item Data flow graph (DFG) match
        \item Simulator feedback via RoboSim~\cite{robo_instruct}
    \end{itemize}
\end{itemize}

The total reward at time step $t$ is defined as the weighted sum of these four components, where $\lambda_{\text{sync}}$, $\lambda_{\text{lint}}$, $\lambda_{\text{KL}}$, and $\lambda_{\text{opt}^i}$ are tunable coefficients that control the relative contribution of each reward component:%
\begin{multline*}
    R_t = \lambda_{\text{sync}} \, R_{\text{sync}}(t) + \lambda_{\text{lint}} \, R_{\text{lint}}(t) \\
    + \lambda_{\text{KL}} \, R_{\text{KL}}(t) + \sum_i\lambda_{\text{opt}^i} \, R_{\text{opt}^i}(t)
\end{multline*}

\subsection{Reward Attribution}

\begin{figure}%
\centering
\includegraphics[width=1.0\linewidth]{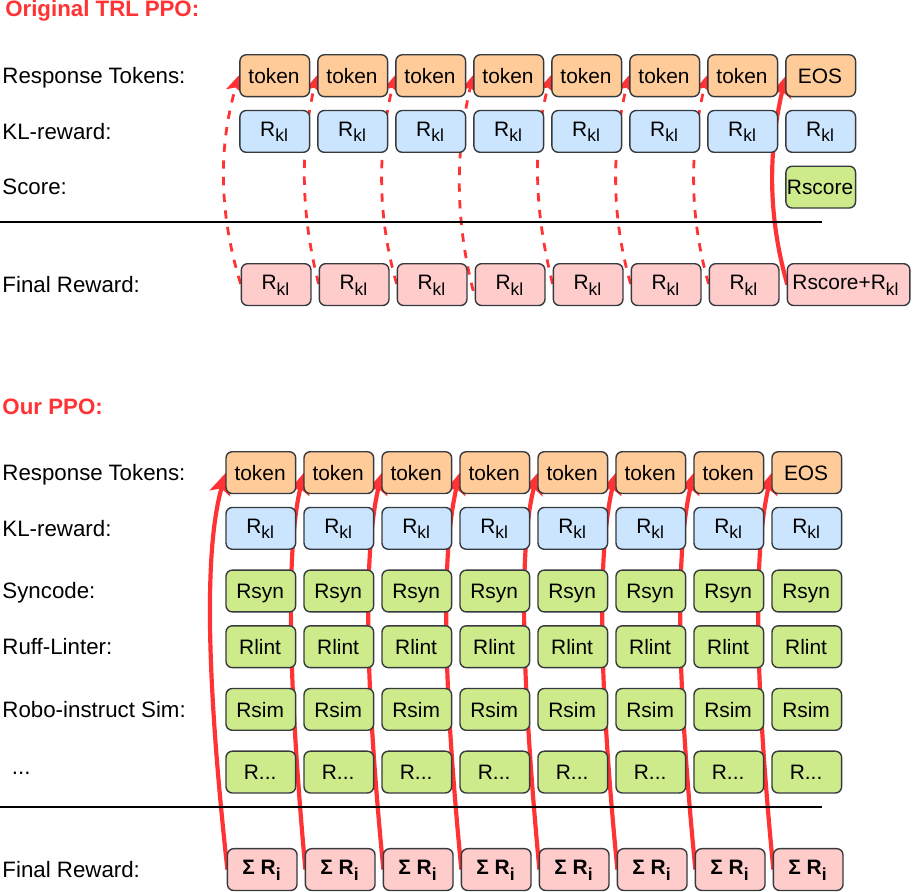}
\caption{Comparison between standard sequence-level rewards (top) and our token-level rewards (bottom).}
\label{fig:token-level-reward}
\end{figure}

To address the reward sparsity problem common in code generation, we introduce a \emph{dense token-level reward}. Figure~\ref{fig:token-level-reward} illustrates the difference to standard implementations, such as the \texttt{trl} PPO algorithm~\cite{trl}. Our PPO variant provides a continuous \emph{sum of all reward components} $\sum R_t$ at every token, whereas \texttt{trl} calculates only the KL divergence per token and aggregates other reward components at the end-of-sequence (EOS) token.

While metrics such as linter checks or simulator feedback are inherently sequence-level (requiring the full code to evaluate), we implement an attribution mechanism for these rewards that identifies the violations or successes in the output text and maps them back to the responsible tokens or token spans. For rewards where precise attribution is not possible, such as unit test outcomes, the reward or penalty is distributed uniformly across the generated code tokens, while non-code tokens receive no reward signal. This creates a dense, localized feedback signal $R_t$ at every time step $t$, significantly accelerating convergence compared to sparse terminal rewards.

\section{Evaluation}

We evaluate the proposed RL framework on two domains: general-purpose Python code generation and robotic program synthesis. In both cases, we train a \texttt{Qwen2.5-Coder-1.5B-Instruct} model~\cite{qwen_coder, qwen_coder_model} and compare its performance before and after fine-tuning. Detailed training configurations for both experimental setups are provided in Appendix~\ref{sec:exps}.

\subsection{General-Purpose Python Code Generation}
\label{sec:eval-mbpp}

\begin{table}%
\centering
\caption{Pass@1 results on MBPP and MBPP+.}
\label{tab:mbpp-results}
\scriptsize
\begin{tabular}{lcc}
\toprule
Model & MBPP & MBPP+ \\
\midrule
Qwen2.5-Coder 1.5B (base) & 0.460 & 0.413 \\
\textbf{Qwen2.5-Coder 1.5B (ours)} & \textbf{0.653} & \textbf{0.556} \\
\midrule
\textit{Absolute Improvement} & \textit{+0.193} & \textit{+0.143} \\
\bottomrule
\end{tabular}
\end{table}

For this experiment, we train on a subset of the OpenCodeInstruct dataset~\cite{opencodeinstruct} and use unit test results and DFG match as additional reward functions. The model is evaluated on MBPP and MBPP+~\cite{mbpp, evalplus} using pass@1. The results are summarized in Table~\ref{tab:mbpp-results}. The RL-refined Qwen2.5-Coder model improves performance from 0.460 to 0.653 on MBPP and from 0.413 to 0.556 on MBPP+, demonstrating substantial gains in correctness and robustness to edge cases. These improvements indicate that the multi-component reward function promotes structurally coherent and semantically consistent program generation beyond surface-level pattern matching.

\subsection{Robotic Program Synthesis}
\label{sec:eval-robotics}

\begin{table}%
\centering
\caption{Task outcomes on \textsc{RoboEval}.}
\label{tab:robotics-outcomes}
\scriptsize
\setlength{\tabcolsep}{3pt}
\begin{tabular}{lcccc}
\toprule
Qwen2.5-Coder model & Python err. & Simulation err. & Completion err. & Success \\
\midrule
1.5B (base) & 77 & 3 & 0 & 0 \\
\textbf{1.5B (ours)} & \textbf{11} & \textbf{28} & \textbf{27} & \textbf{14} \\
\midrule
7B (base) & 5 & 19 & 26 & 30 \\
7B (\textsc{Robo-Instruct}) & 9 & 16 & 24 & 31 \\
\bottomrule
\end{tabular}
\end{table}

For his experiment, we train on the \textsc{Robo-Instruct} dataset~\cite{robo_instruct_dataset} and use the simulator feedback by RoboSim~\cite{robo_instruct} as additional reward function. The model is evaluated on the \textsc{RoboEval} benchmark~\cite{mobile_robots}. Table~\ref{tab:robotics-outcomes} summarizes the task outcomes. A \emph{Python error} is a (usually syntactic) error that prevents compiling and executing the code. A \emph{simulation error} is raised by RoboSim if executing the generated program is infeasible because it violates the physical constraints of the simulation environment. A \emph{completion error} indicates that the program was executable but did not solve the task.

The base model fails primarily at the Python-level, producing errors in 77 out of 80 tasks, meaning that only 3 tasks reach the simulator, none of which is executable. After RL fine-tuning, Python-level errors are reduced to 11, and 14 tasks are correctly solved. Overall, we see a shift form 100\% non-executable code to 51\% code that can successfully be executed in the simulator. These improvements indicate that our reward composition enables the model to generate environment-aware programs.

For comparison, we also show the outcomes for the Qwen2.5-Coder 7B model trained by Hu et~al.~\cite{robo_instruct} with \textsc{Robo-Instruct} and the corresponding base model. While our fine-tuned 1.5B model does not achieve performance competitive with the larger 7B models, the relative performance gain through our RL approach is much larger than that of \textsc{Robo-Instruct}, narrowing the gap to substantially larger models.

\section{Related Work}

Recent advances in large language models have significantly improved automated code generation, enabling applications such as program synthesis, code completion, translation, and program repair. To improve functional correctness beyond supervised learning, reinforcement learning has emerged as a key paradigm for fine-tuning code generation models using execution-based feedback. Representative approaches such as CodeRL~\cite{coderl}, PPO-Coder~\cite{ppocoder}, and CoTran~\cite{cotran} formulate code generation as a sequential decision-making problem and optimize policies using rewards derived from program execution. These methods demonstrate improvements in syntactic validity and functional correctness, but remain limited by sparse, usually sequence-level reward signals and a lack of awareness of domain-specific constraints. RLVR~\cite{qwen_coder, deepseekcoder}, while effective, has similar limitations. Distillation-based methods, such as SDPO~\cite{sdpo}, provide more fine-grained feedback but are computationally expensive and only work well with very large models.

In parallel, LLMs have been increasingly applied to robotic systems, enabling natural language instructions to be translated into executable robot programs~\cite{llm_for_robotic_control_tasks, llm_for_robot_coding_edjucation, robo_script}. Existing approaches typically follow one of three paradigms: end-to-end fine-tuning~\cite{rt1, palme}, prompt-based task decomposition~\cite{inner_monologue, llm_planner}, and high-level program generation~\cite{sayplan, code_as_policy}. Among these, high-level program generation---where LLMs write code in a high-level programming language that invokes robot control primitives---has shown the strongest generalization capability. A notable example is Code-as-Policies~\cite{code_as_policy}, which demonstrates that LLMs can perform spatial reasoning and translate natural language instructions into parameterized robot behaviors.

Despite these advances, robotic code generation introduces additional challenges not addressed by general-purpose RL-based code generation methods. These include safety-critical execution, sensitivity to instruction variations, and the lack of robust evaluation frameworks that go beyond simple input-output correctness~\cite{mobile_robots}. Existing benchmarks often fail to capture whether generated programs correctly execute full task sequences in real or simulated environments.

Our approach differs from prior work by integrating multi-level, environment-aware execution feedback directly into the RL optimization loop. Rather than relying solely on sparse sequence-level rewards, we introduced dense token-level reward attribution within a PPO-based framework. This integration enables iterative alignment of code-generating LLMs toward both functional correctness and executable robotic behavior, addressing gaps left by earlier RL methods.

\section{Conclusion}

This work investigated whether reinforcement learning with structured, execution-aware rewards can improve LLM-based code generation across different domains. We introduced a unified fine-tuning framework based on PPO that combines syntactic, functional, security, and simulator-based feedback with token-level reward mapping for improved credit assignment. Experiments on MBPP, MBPP+, and \textsc{RoboEval} demonstrated substantial improvements in both functional correctness and simulator-executable code generation. In robotics, RL reduced execution failures by 51\% and enabled a transition from largely non-executable outputs to environment-aware program generation.

Future work should improve alignment between task instructions and generated robotic behavior, investigate normalization strategies to reduce length bias in rewards, design more general execution-based reward mechanisms beyond assertion-style unit tests, and explore reward functions for new domains. Additionally, systematic ablation studies are needed to better understand the contribution of token-level rewards and alternative credit assignment strategies. Lastly, it should be investigated to what degree larger, more capable LLMs can benefit from the proposed fine-tuning approach.

Overall, the results show that structured reinforcement learning is a practical and extensible approach for aligning code-generating language models with program correctness and environment-aware execution.

\section*{Acknowledgements}

This work was funded by the Deutsche Forschungsgemeinschaft (DFG, German Research Foundation) under Germany's Excellence Strategy (EXC-3057/1 ``Reasonable
Artificial Intelligence'', Project No. 533677015), by the National Research Center for Applied Cybersecurity ATHENE within the project ``Foundational Models for Secure Software Development'', and by the LOEWE initiative (Hesse, Germany) [LOEWE/4a//519/05/00.002(0013)/95]. We gratefully acknowledge support from the hessian.AI Service Center (funded by the Federal Ministry of Research, Technology and Space, BMFTR, grant no. 16IS22091) and the hessian.AI Innovation Lab (funded by the Hessian Ministry for Digital Strategy and Innovation, grant no. S-DIW04/0013/003).

\section*{Impact Statement}

This paper presents work whose goal is to advance the field of Machine Learning. There are many potential societal consequences of our work, none which we feel must be specifically highlighted here.

\bibliography{bibliography/reference}
\bibliographystyle{meta/icml2026}

\newpage
\appendix
\onecolumn

\section{Additional Experimental Details}
\label{sec:exps}

\subsection{Experiment I: General-Purpose Python Generation}
\label{sec:exp-general-code}

The first experiment evaluates the framework's applicability to general-purpose Python code generation from natural language instructions. We investigate whether combining complementary reward signals---covering syntactic correctness, semantic structure, and functional correctness---enhances the overall code quality and correctness of LLMs. Specifically, the reward function integrates \textsc{SynCode}, Ruff-Linter feedback, Data Flow Graph (DFG) matching, and unit-test–based functional correctness signal.

\paragraph{Task Environment}

For this experiment, the policy is fine-tuned on a subset of the OpenCodeInstruct dataset~\cite{opencodeinstruct}, which contains approximately 5 million high-quality instruction–code pairs. To ensure a focused and reproducible training session, we randomly sampled 5,000 examples from the dataset. A fixed random seed is used to maintain consistency across experimental runs and ensure that the training distribution remains identical during hyperparameter tuning. We performed limited manual tuning on a held-out subset prior to final training.

\paragraph{Training Configuration}

 We fine-tune the policy for four PPO epochs using the composite reward formulation introduced in Section~\ref{sec:composite-reward}. The pass@1 unit test reward serves as the primary optimization signal for functional correctness, while syntactic, and structural semantic rewards act as auxiliary shaping signals that guide generation toward valid and well-structured code. Reward weights and PPO hyperparameters are summarized in Table~\ref{tab:hyperparams-python}.

\begin{table}[htb!]
\centering
\caption{Hyperparameters and reward weights for general-purpose Python generation.}
\label{tab:hyperparams-python}
\begin{tabular}{lclc}
\toprule
PPO hyperparameter & Value & Reward component & Weight ($\lambda$) \\ 
\midrule
Learning rate & $3 \times 10^{-6}$ & Syntax ($R_{\text{sync}}$) & 0.1 \\
KL coefficient ($\beta$) & 0.1 & Linter ($R_{\text{lint}}$) & 0.1 \\
Mini-batch size & 8 & DFG semantic ($R_{\text{dfg}}$) & 0.1 \\
PPO epochs & 4 & Pass@1 unit-test reward ($R_{\text{pass@1}}$) & 0.7 \\
Clip range & 0.2 & & \\ 
\bottomrule
\end{tabular}
\end{table}

\paragraph{Hardware and Training Framework:} 
Fine-tuning is performed on two NVIDIA A100 GPUs with 80 GB of VRAM each. We use DeepSpeed together with Hugging~Face Accelerate to efficiently manage memory, enable mixed-precision training, and accelerate distributed optimization across GPUs.

\paragraph{Model Architecture and Parameter-Efficient Fine-Tuning:}
To ensure training efficiency while preserving reasoning capabilities, we apply Parameter-Efficient Fine-Tuning (PEFT) using Low-Rank Adaptation (LoRA)~\cite{lora} on the \texttt{Qwen2.5-Coder-1.5B-Instruct}~\cite{qwen_coder, qwen_coder_model} foundation model. This choice aligns with the framework's design goal of enabling rapid adaptation across domains without modifying the underlying architecture.

\paragraph{Inference Configuration:}
Unless stated otherwise, all evaluations use near-deterministic decoding with a temperature of 0.01, top-$p$ sampling of 0.95, and a maximum generation length of 1024 tokens. This controlled decoding regime minimizes stochastic variation, ensuring that observed performance differences primarily reflect policy improvements rather than sampling noise. Generated responses are processed using a regex-based extractor to isolate the Python function from accompanying natural language explanations. The extracted code is then submitted to the EvalPlus~\cite{evalplus} execution sandbox for verification, ensuring consistent and reproducible evaluation of functional correctness.

\subsection{Experiment~II: Robotic Task Alignment}
\label{sec:exp-robotics}

This experiment evaluates the proposed framework in a domain where code correctness depends on physical feasibility and action sequencing. We assess whether integrating simulator-based feedback into the RL loop enables the model to generate robot programs that are executable under realistic constraints.

\paragraph{Task Environment}

We conduct the robotics experiments on the \textsc{Robo-Instruct} dataset~\cite{robo_instruct_dataset}, which provides approximately 5,000 instruction--program pairs grounded in a simulated service robot environment. Each task requires generating Python code that interacts with a fixed set of high-level robot APIs and can be executed in simulation.

The simulator enables environment-aware verification by detecting infeasible actions such as collisions, invalid object manipulation, or unreachable navigation goals, and by monitoring task-level state transitions during program execution. This simulation-based feedback is used both during training as a task-specific reward signal and during evaluation through the \textsc{RoboEval}~\cite{mobile_robots} benchmark.

\paragraph{Training Configuration}

Training uses the same PPO architecture and optimization pipeline as Experiment I. The hardware and training framework are also identical. In this experiment, the simulator-based task is given the highest weight compared to the syntax and linter components, prioritizing functional correctness over purely surface-level code quality. The complete set of PPO hyperparameters and reward weights is summarized in Table~\ref{tab:hyperparams}.

\begin{table}[htb!]
\centering
\caption{Hyperparameters and reward weights for robotics code generation.}
\label{tab:hyperparams}
\begin{tabular}{lclc}
\toprule
PPO hyperparameter & Value & Reward component & Weight ($\lambda$) \\ 
\midrule
Learning rate & $1 \times 10^{-6}$ & Syntax ($R_{\text{sync}}$) & 0.1 \\
KL coefficient ($\beta$) & 0.9 & Linter ($R_{\text{lint}}$) & 0.1 \\
Mini-batch size & 4 & Simulation ($R_{\text{sim}}$) & 0.8 \\
PPO epochs & 4 & & \\
Clip range & 0.2 & & \\ 
\bottomrule
\end{tabular}
\end{table}

\section{Additional Results}

\subsection{Experiment I: General-Purpose Python Code Generation}
\label{sec:res-general}

Experiment~I evaluates whether the proposed multi-component RL framework improves functional correctness and structural quality in general-purpose Python code. Unlike grounded robotic planning tasks, where correctness is constrained by physical feasibility, this experiment measures static program correctness using unit-test-based evaluation.

We report pass@1 performance for both the base Qwen2.5-Coder model and the RL-refined policy. Results are summarized in Table~\ref{tab:mbpp-results}. Performance is assessed using the MBPP benchmark as well as its extended variant MBPP+ (EvalPlus)~\cite{mbpp, evalplus}, which introduces additional test cases to expose brittle logic and edge-case failures.

The RL-refined policy achieves substantial improvement over the base model, with a 19.3 percentage point increase on MBPP. Gains persist on MBPP+, where a 14.3 percentage point improvement demonstrates enhanced robustness to edge cases rather than overfitting to canonical solutions.

These results suggest that the multi-component reward function encourages the generation of structurally coherent programs with robust control flow and data dependencies, beyond surface-level pattern matching.

\subsection{Experiment II: Robotic Task Alignment}
\label{sec:res-robotics}

Experiment~II evaluates whether simulator-aware reward shaping enables a compact language model to generate robot programs that are syntactically valid, physically feasible, and executable under environment-specific constraints.

Performance is measured using pass@1 on the \textsc{RoboEval} benchmark~\cite{mobile_robots}, with a breakdown of failure modes to characterize behavioral changes under RL.

Table~\ref{tab:robotics-outcomes} summarizes outcomes before and after fine-tuning for the Qwen2.5-Coder 1.5B model trained with our RL framework as well as the Qwen2.5-Coder 7B model trained with \textsc{Robo-Instruct} by Hu et~al.~\cite{robo_instruct}. Each model is evaluated on 80 task instances. The different outcomes are explained in Section~\ref{sec:eval-robotics}.

\section{Training Algorithm}
\label{sec:training-algorithm}

Alg.~\ref{alg:ppo-syncode} summarizes the full training loop used in our framework. For each input–output pair $(x, y)$ from the dataset, the policy model generates one candidate \textit{response} $\hat{y} \sim \pi_\theta(\cdot|x)$. Each response is then evaluated using the multi-component reward system introduced in Section~\ref{sec:composite-reward}. Using the aggregated reward, we compute advantage estimates and update both the policy and the value model via the clipped PPO objective. This iterative procedure gradually steers the model toward producing code that is syntactically valid, semantically meaningful, and well aligned with the target task.

\begin{algorithm}[p]
\caption{Proximal policy optimization for code generation with multi-component token-wise rewards.}
\label{alg:ppo-syncode}
\begin{algorithmic}[1]

\REQUIRE dataset $\mathcal{D}$, policy $\pi_\theta$, reference model $\pi_{\text{ref}}$, value model $V_\phi$
\REQUIRE Scoring: linter $f_{\text{lint}}$, \textsc{SynCode} $f_{\text{syn}}$, user-defined task reward functions $\mathcal{R}_{\text{task}} = \{r_1, r_2, \dots, r_m\}$
\REQUIRE Hyperparams: $\beta$ (KL), $\gamma, \lambda$ (GAE), $\epsilon$ (Clip), Coefficients $\{c_1, c_2, \alpha_k\}$
\STATE Initialize global step $t \gets 0$

\WHILE{training not finished}
    \STATE \textit{// --- Rollout Phase ---}
    \STATE Sample batch of queries $x \sim \mathcal{D}$
    \STATE Generate responses $y$ and \textsc{SynCode} scores $s_{\text{syn}}$ via $\pi_\theta(\cdot|x)$ guided by $f_{\text{syn}}$
    \STATE Compute log-probs $\log \pi_\theta(y|x)$ and $\log \pi_{\text{ref}}(y|x)$

    \STATE \textit{// --- Evaluation Phase ---}
    \FOR{each sample $(x_i, y_i)$ in batch}
        \STATE $V[i] \gets V_\phi(x_i, y_i)$ \COMMENT{// critic value estimate}
        \STATE $s_{\text{lint}}[i] \gets f_{\text{lint}}(y_i)$ \COMMENT{// linter score}
        \STATE Compute sample KL divergence: $D_{\text{KL}}[i] \gets \log \pi_\theta(y_i|x_i) - \log \pi_{\text{ref}}(y_i|x_i)$
        \STATE Combine weighted feedback: $S_{\text{raw}}[i] \gets c_1 \cdot s_{\text{syn}}[i] + c_2 \cdot s_{\text{lint}}[i] + \sum_{k=1}^m \alpha_k r_k(y_i)$
        \STATE Total reward per sample: $R[i] \gets S_{\text{raw}}[i] - \beta \cdot D_{\text{KL}}[i]$
    \ENDFOR
    \STATE Compute advantages $A$ and returns $G$ using GAE($\gamma, \lambda$) on $R$ and $V$

    \STATE \textit{// --- Optimization Phase ---}
    \FOR{epoch $k = 1$ \TO $K_{\text{epochs}}$}
        \STATE Shuffle and split data into mini-batches $\mathcal{M}$
        \FOR{each mini-batch $B \in \mathcal{M}$}
            \STATE Compute policy ratio: $\rho_t \gets \exp(\log \pi_\theta(y|x) - \log \pi_{\theta_{\text{old}}}(y|x))$
            \STATE Update $\pi_\theta$ by maximizing PPO objective:
            \[
                \mathbb{E} \left[ \min\left( \rho_t A_t,\ \text{clip}(\rho_t, 1-\epsilon, 1+\epsilon) A_t \right) \right]
            \]
            \STATE Update value model $V_\phi$ to minimize mean squared error
        \ENDFOR
    \ENDFOR
\ENDWHILE

\end{algorithmic}
\end{algorithm}

\end{document}